\documentclass{article}
\vspace{-4ex}
\date{}

\usepackage{natbib}
\setcitestyle{numbers}

\usepackage[a4paper, total={7in, 10in}]{geometry}
\usepackage[utf8]{inputenc} % allow utf-8 input
\usepackage[T1]{fontenc}    % use 8-bit T1 fonts
\usepackage{hyperref}       % hyperlinks
\usepackage{url}            % simple URL typesetting
\usepackage{booktabs}       % professional-quality tables
\usepackage{amsfonts}       % blackboard math symbols
\usepackage{nicefrac}       % compact symbols for 1/2, etc.
\usepackage{microtype}      % microtypography
\usepackage{amsmath}
\usepackage[ruled,vlined]{algorithm2e}
\usepackage{multirow}
\usepackage{array}
\usepackage{booktabs}
\usepackage{graphicx}
\usepackage{tabularx}
\usepackage{amsmath}
\usepackage{ulem}

\usepackage{tikz,graphics,color,fullpage,float,epsf,caption,subcaption}

\setcitestyle{square}

\usepackage{color}

\usepackage{authblk}
\usepackage{color}

\title{Reprogramming Pretrained Language Models for Protein Sequence Representation Learning}

\author[1]{Ria Vinod
}
\author[2]{Pin-Yu Chen}
\author[2]{Payel Das}
\affil[1]{Department of Computational and Molecular Biology, Brown University}
\affil[2]{IBM Research}
\begin{document}

\maketitle

\begin{abstract}
    %Machine Learning-guided solutions for molecular learning tasks on biochemical data have made significant headway in recent years. However, success in scientific discovery tasks is limited by the accessibility of well-defined and labeled in-domain data. To tackle the low-data constraint, recent adaptions of deep learning models pretrained on millions of protein sequences have shown promise; however, the construction of such domain-specific large-scale model is computationally expensive. Here, we propose Representation Learning via Dictionary Learning (R2DL), an end-to-end representation learning framework in which we reprogram deep  models for alternate tasks that can perform well in biochemical data domains with significantly fewer training samples. R2DL reprograms a pretrained English language model to learn the embeddings of protein sequences, by learning a sparse linear mapping between English and protein sequence vocabulary embeddings. Our method can attain high data efficiency because the number of trainable parameters is orders less than the parameters of the pretrained model. Consequently, our model can attain better accuracy and significantly improve the data efficiency by up to 105 times over the baselines set by pretraining and standard supervised methods. To this end, we reprogram a transformer and benchmark on a biologically relevant set of protein physicochemical prediction tasks (secondary structure, stability, homology, stability) and a biomedically relevant set of protein function-related tasks (antimicrobial, toxicity, antibody affinity). 
    Machine Learning-guided solutions for protein learning tasks have made significant headway in recent years. However, success in scientific discovery tasks is limited by the accessibility of well-defined and labeled in-domain data. To tackle the low-data constraint, recent adaptions of deep learning models pretrained on millions of protein sequences have shown promise; however, the construction of such domain-specific large-scale model is computationally expensive. Here, we propose Representation Learning via Dictionary Learning (R2DL), an end-to-end representation learning framework in which we reprogram deep  models for alternate-domain tasks that can perform well on protein property prediction with significantly fewer training samples. R2DL reprograms a pretrained English language model to learn the embeddings of protein sequences, by learning a sparse linear mapping between English and protein sequence vocabulary embeddings. Our model can attain better accuracy and significantly improve the data efficiency by up to $10^5$ times over the baselines set by pretrained and standard supervised methods. To this end, we reprogram an off-the-shelf pre-trained English language transformer and benchmark it on a  set of protein  physicochemical prediction tasks (secondary structure, stability, homology, stability) as well as on a biomedically relevant set of protein function prediction tasks (antimicrobial, toxicity, antibody affinity).
    
\end{abstract}

\section*{Introduction}

Recent advances in artificial intelligence (AI), particularly in deep learning, have led to major innovations and advances in many scientific domains, including biology. These deep learning models aim to learn a highly accurate and compressed representation of the biological system, which then can be employed for a range of tasks. There has been notable success across a range of tasks, from high-quality protein structure prediction from protein sequences \cite{jumper2021highly, baek2021accurate}, accurate prediction of protein properties, to enabling novel and functional peptide discoveries \cite{das2021accelerated, stokes2020deep}. Many of these advances rely on developing deep learning models \cite{jumper2021highly, meier2021language, rao2019evaluating} which are trained from scratch on massive amounts (on the order of billions of tokens) of data. However, labeled data in biology is scarce and sparse, which is also the case for many other real-world scenarios in the scientific domain. In the biological domain, label annotation can involve biological assays, high resolution imaging and spectroscopy, which are all costly and time consuming processes.

The technique of pretraining deep learning models was proposed to address this issue. Pretraining methods leverage large amounts of sequence data and can learn to encode features that can explain the variance seen in sequences across biological task-specific training samples. In the context of protein sequences, pretraining has enabled meaningful density modelling across protein functions, structures, and families \cite{https://doi.org/10.48550/arxiv.2206.13517}. In this work, we reference two types of pretraining methods: (i) unsupervised pretraining, where all data is unlabeled, and (ii) self-supervised pretraining, where a model learns to assign labels to its unlabeled data. Large models then pretrain on massive amounts of unlabeled data, specifically biological sequences, which are available at scale. Once pretrained, these foundation models (FMs) \cite {bommasani2021opportunities} are finetuned on smaller amounts of labeled data, which correspond to a specific downstream  task. Interestingly, for the  large-scale models pretrained on protein sequences, biological structure and function seem to emerge in the learned protein representation, even though such information was not included in model training \cite{meier2021language}. 

Though highly powerful, the training of those domain-specific foundation models from scratch is  highly resource-intensive \cite{yuan2022decentralized}. For example, one training run of BERT (the language model considered in this work) learns 110 million parameters, costs up to \$13,000 USD and takes 64 days (without parallelized computing) and results in 0.7 tons of carbon emissions \cite{devlin2018bert}. A single training run of another popular language model, the T5 transformer, learns 11 billion parameters, costs up to \$1.3 million USD, takes 20 days, and results in 47 tons of carbon emissions \cite{patterson2021carbon, raffel2020exploring}. Such pretrained language models and size variants are abundantly available with the advent of models libraries (e.g., Hugging Face \cite{https://doi.org/10.48550/arxiv.1910.03771}) which host pretrained models and datasets. The scale of data, compute, and financial resources required to train these models is not only available to a limited number of researchers, but is also infeasible for applications with limited labeled data. However, in the scientific domain, we still aim to train models with similar representational capacity and predictive performance. To this end, we propose a lightweight, and more accurate alternative method to large-scale pretraining. %This method is also robust in limited data regimes. 
Specifically, we introduce a method to reprogram an existing foundation model of high capacity that is trained on data from a different domain. This situation calls for innovations in cross-domain transfer learning, which is largely unexplored, particularly in scientific domains.

%For example, one training run of GPT3-175B takes 20 days %3.6K Petaflops-days, 
%and costs approximately \$5 million USD \cite{brown2020language} and results in 552 tons %of carbon emissions \cite{patterson2021carbon}. 

% mention reprogramming, biological reprogramming, robustness vs training efficiency

\begin{figure}[t]
\centering
\includegraphics[scale=0.8]{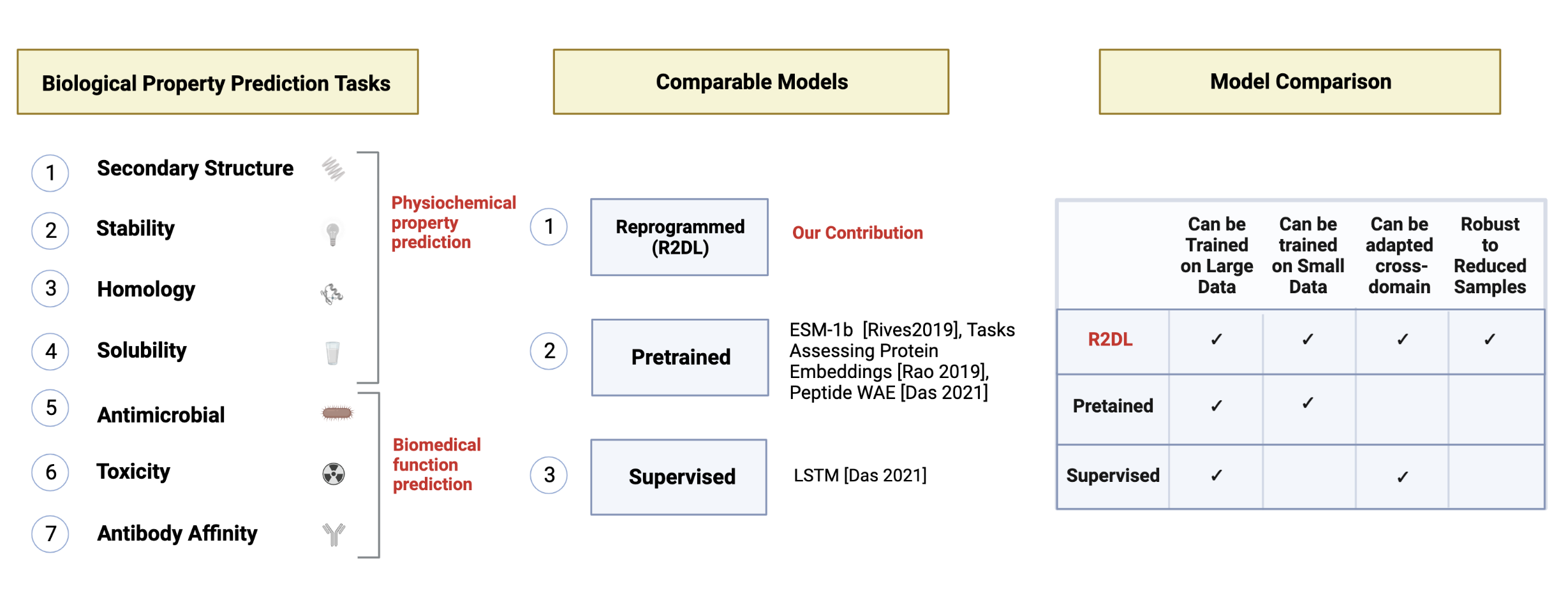}
\caption{\textbf{Left:} Descriptions of considered predictive tasks. %center: comparable models; right: model capabilities. 
We select the set of  physicochemical property prediction tasks from the well-studied domains in \cite{rao2019evaluating}, and the biomedical function prediction tasks from works with biomedically relevant small-szied labeled datasets \cite{das2021accelerated, makowski2022co}. \textbf{Center:} We compare R2DL to pretraining and standard supervised training methods. We refer to supervised methods as standard supervised classifiers that are trained from scratch from labeled data alone. Depending on how labeled and unlabeled data are used in pretraining, we consider pretraining to constitute unsupervised/supervised pretraining. \textbf{Right:} The comparative table shows the broad adaptability of the R2DL framework. In comparison to existing gold standard methods, R2DL is has a broader utility across different domains, sizes of training datasets, and data efficiency. We categorize supervised methods as cross-domain adaptable, through various domain adaptation and transfer learning techniques \cite{zhuang2020comprehensive}. }
\end{figure}

%\RV{cite} before warm start
One known fact is that biological sequences are similar to natural language, as they also contain long-range dependencies and follow Zipf’s law \cite{newman2005power}. These sequences and their associated dependencies are crucial for determining their structural and functional properties. Such similarity has motivated the use of deep learning architectures and mechanisms that are widely used in natural language processing (NLP) to build protein sequence models from scratch. In this work, we explore an alternative \textit{warm-start} paradigm, i.e. how to effectively and efficiently reprogram an existing, fully-trained large English language model to learn a meaningful (i.e., biomedically relevant) representation of protein sequences. The goal is to create a more carbon-friendly, resource-efficient, and broadly accessible framework to motivate different scientific domains toward democratizing the representation power of large AI models. This \textit{warm-start} paradigm is defined by the framework's ability to achieve the performance of transformers that are pretrained on billions of tokens, with a lighter-weight training procedure that is similar to that of a standard supervised classifier trained from scratch. In particular, we consider highly specific biological and biomedical protein sequence datasets (illustrated in Figure 1) which have much fewer samples than standard supervised language task datasets. Reprogramming thus provides a more data and resource-efficient approach to developing models to achieve deep representational capacity and performance for downstream protein tasks. Reprogramming has been previously explored in the language domain as a sub-problem of transfer learning \cite{chen2022model}. \cite{neekhara2018adversarial} explored reprogramming language models for alternate text classification tasks, \cite{yang2021voice2series} reprogrammed acoustic models for time series classification, \cite{elsayed2018adversarial} reprogrammed ImageNet classification models for alternate image classification tasks. However, none of these methods investigate mappings between domains that require a very high representational capacity (from natural language to biological sequence), which is the setting we require in the protein sequence domain.

Toward this goal, we introduce R2DL (Representation Reprogramming via Dictionary Learning),  a novel cross-domain transfer learning framework to reprogram an existing pretrained large-scale deep-learning model of the English language, namely a English BERT model \cite{devlin2018bert}, to learn and predict physicochemical and biomedical properties of protein sequences. To the best of our knowledge, our work remains the first work to address reprogramming in any biological, and more broadly, scientific domain. In Figure 1, we illustrate the set of protein  physicochemical and functional property prediction tasks we consider, as well as the baseline methods against which we compare R2DL performance to, and a brief description of R2DL's advantages compared to these existing methods. We test the reprogrammed model for a range of biomedically relevant downstream physicochemical property, structure, and function prediction tasks, which include  prediction of secondary structure, homology, mutational stability, solubility, as well as antimicrobial nature, toxicity, and antibody affinity of proteins. Each of these tasks involves learning on datasets which are limited to a few thousands of labeled samples,  at least an order of magnitude smaller needed to train a foundation model or a large language model \cite{https://doi.org/10.48550/arxiv.2108.07258}. R2DL uses dictionary learning, a machine learning framework that finds the optimal sparse linear mapping between the English vocabulary embeddings and the amino acid embeddings. To do so, a protein property prediction task-specific loss is used to learn the optimal parameters of the reprogrammed model. We train R2DL in a supervised setting with the downstream protein prediction task datasets that are labeled and small in size (illustrated in Figure 1). R2DL demonstrates consistent performance improvement from existing baselines across seven different physicochemical (e.g., up to 11\% in stability), structural,  and functional property prediction (e.g., up to 3\% in toxicity) tasks of proteins. We estimate R2DL to be over 105 times more data-efficient than existing pretraining methods. We further demonstrate the performance robustness of R2DL when trained on a reduced size version of the supervised protein datasets. In addition, we show that that R2DL learns to encode physicochemical and biomedical properties in the learned representations, even in a limited data scenarios. This work thus blazes a path toward efficient and large-scale adaptation of existing foundation models toward different real-world learning tasks and accelerates scientific discovery, which naturally involves learning from limited real-world data.

\section*{Results}

Figure 2 illustrates the proposed Representation Reprogramming via Dictionary Learning (R2DL) framework, which learns to embed a protein sequence dataset of interest by training on the  representations of a transformer that is pretrained on an English text corpus. A one-to-one label mapping function is assigned for each downstream protein prediction task for cross-domain machine learning, and a class label or a regression value is predicted using R2DL for each protein sequence during testing. Below we discuss details of the general framework (tasks described in Figure 1). 

\begin{figure}[t]
\centering
\includegraphics[scale=0.8]{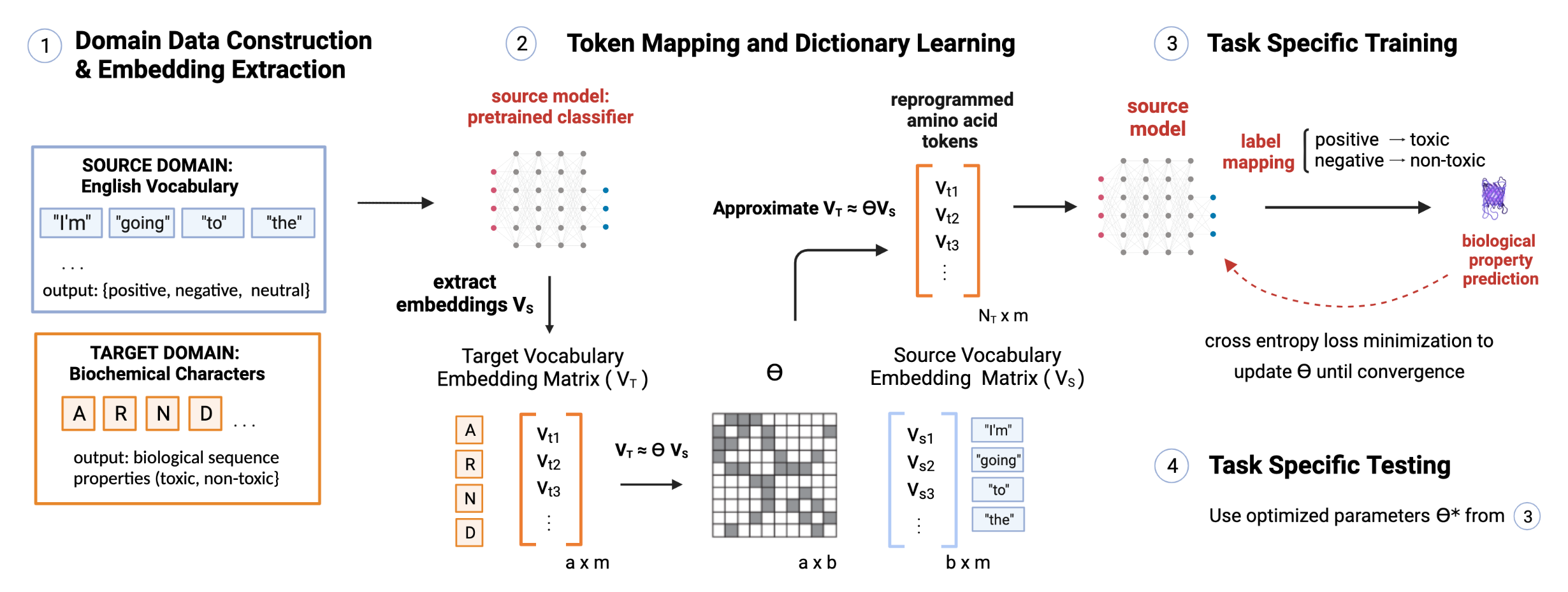}
\caption{System illustration of the Representation Reprogramming via Dictionary Learning (R2DL) framework. In Step 1, R2DL loads a pretrained language model (source), obtains the source vocabulary embeddings, and specifies the protein tokens (target). In Step 2, R2DL learns a sparse linear mapping between the source and target embeddings via dictionary learning, to represent a target token embedding as a sparse linear combination of source token embeddings. In Step 3, the system maps the source task labels (e.g., positive/negative sentiments) to target task labels (e.g., toxic/non-toxic proteins) and optimizes the embedding mapping parameters based on the task-specific loss evaluation on a given protein sequence dataset. Finally, in Step 4 the reprogrammed model is deployed for the test-time evaluation.}
\end{figure}

\subsection*{R2DL Framework Formulation}
The R2DL objective is to reprogram a source model (pretrained language model) to be able to correctly classify, or predict the regression values of, protein sequences (for a target prediction task). We use pretrained instances of BERT, a bidirectional transformer (termed the source model), which has been finetuned separately for different language tasks (e.g., sentiment classification, named entity recognition) \cite{devlin2018bert, jiao2019tinybert}. 
%R2DL assumes access to the gradients of the source model and this approach works in a semi-black box setting given that we access but do not modify the internal architecture. 
For a protein sequence classification task, we use the source model trained on a language task for which there are $n$ sentence output classes (e.g., positive and negative for senitiment classification), and $n$ protein sequence classes (e.g., toxic, non-toxic). The output-label mapping $h$ is then a simple 1-1 correspondence between the source task labels and the target task labels (e.g.,  positive $\rightarrow$ toxic and negative $\rightarrow$ non-toxic). For a regression task, R2DL uses a mapping between the regression values in protein sequence feature space and the classification probability values in the source model embedding space. It does so by learning optimal thresholds of regression values that map to the source model class labels.

The input data of the source English language model is tokenized at the word level. These tokens form the atoms for our dictionary representation of $V_S$, a matrix with its rows corresponding to embedding vectors of source tokens. The input data to the target task, protein sequences, are tokenized on a character level with only 20 distinct tokens (corresponding to the set of 20 discrete natural amino acid characters). R2DL obtains $V_S$ from the learned embeddings of the source model and learns to represent $V_T$, the matrix of the target token embedding, as a weighted combination of the English token embeddings. We propose token reprogramming by approximating a linear mapping between $V_S$ and $V_T$. That is, we aim to find a transformation of the latent representation of the protein sequences, such that it can be embedded in the pretrained language model's latent space and enable R2DL to leverage these re-embedded tokens for learning. Specifically, we learn the linear map $\Theta$ by approximating a dictionary using a k-SVD solver \cite{k-SVD}. That is, we want to approximate $V_T = \Theta V_S$. The k-SVD solver guarantees a task-specific level of sparsity in the coefficients when linearly combining English token embeddings to represent a protein sequence token embedding. In other words, it helps select $k$ English tokens and use their linearly combined embeddings as the embedding of a target token. Additionally, with a one-to-one label mapping function of the protein sequence label to the English text label, we are able to use the pretrained language model for inference on the embedded protein dataset, $V_T$. We thus design an end-to-end reprogramming framework for any arbitrary protein sequence classification or regression task.

\subsection*{R2DL Training and Optimization Procedure}

We are given a pretrained classifier, $\mathbf{C}$  (which has been pretrained on a source-task dataset with source tokens denoted by $\{v_{Si}\}^{|V_S|}_{i=1}$) and a target-task dataset with target tokes denoted by $\{V_{Tj}\}^{|V_T|}_{j=1}$. The embedding matrices are $V_S$ and $V_T$ respectively. We can encode an output label mapping function translating between source and target labels. In Figure 2, we show how R2DL aims to find a linear mapping function $\Theta$ that learns the optimal coefficients for our atoms in $V_T$ to be represented as a sparse encoding of the dictionary $V_S$ such that $V_T = \Theta V_S $. The map $\Theta$ is used to reprogram $\mathbf{C}$ to be able to correctly classify the protein sequences through the transformation $h(\mathbf{C}(\theta_t,t))$ where $t$ is a protein sequence from a protein task and $\theta_t$ is the linear weights associated with the protein sequence $t$ in $\Theta$. We note that for each of the downstream protein property prediction task, R2DL only trains a corresponding token mapping function $\Theta$ while keeping the pretrained classifier $\mathbf{C}$ intact. Therefore, the number of trainable parameters in R2DL is simply the size of the matrix $\Theta$, which is usually much smaller compared to the number of parameters in the pretrained deep neural network classifier $\mathbf{C}$.
%While we do not make any modification the parameters or architecture of $\mathbf{C}$, we assume access to the gradient $\nabla_{x} \Theta$ for loss evaluation during training. 
%To reprogram the pretrained classifier, we use an embedding mapping $f_\theta : s_i \longmapsto t_i$ where $s_i \in V_S$ and $t_i \in V_T$. Dimension of the input space of $V_S$ and $V_T$ is $|V_S|$, and $|V_T|$ respectively, where $|V_T| \ll |V_S|$. $f_\theta$ \RV{fix} is parametrized by $\theta \in \mathbf{R}^{|V_S| \times |V_T|}$, which represents the coefficients of the atoms in $V_S$ such that $V_T = \Theta V_S $. The observation of $|V_T| \ll |V_S|$ requires that our mapping has high representational capacity, so we encode a sparse representation of $V_T$, to extract relevant features from the source vocabulary embeddings $\{V_S\}^{|V_S|}_{i=1}$ for the alternate task.
To approximate the dictionary, we use a k-SVD solver to optimize over the cross entropy loss for updates to $\Theta$. We then apply the assigned label mapping $h$ for protein classification tasks, or thresholding for regression tasks, and train the mapping function $\Theta$ using gradient-based optimization evaluated on the task-specific cross-entropy loss. Details for R2DL training procedure are given in the Method section.

\subsection*{Benchmark Tasks and Evaluation}
%\subsection*{Reprogramming Language Models for Protein Property Classification And Regression Tasks}

%In Figure 1, we illustrate the end-to-end R2DL framework that can be repurposed for any protein predictive task by changing the choice of the source model and label mapping functions. Typical amino acid representations of biological molecules or widely used SMILES representation of chemical molecules have fewer distinct tokens than that of language data. The significant reduction in dimensionality of the embedded space of English data to molecule data makes finding a mapping non-trivial. \cite{bengio_rep} demonstrates results that that representation learning algorithms have an advantage in transfer learning methods as they capture features relevant to alternate tasks. We requires a mapping with high representational capacity. Work in \cite{goodfellow_inv_networks} demonstrates that a sparse encoding is distributed and highly expressive: we can represent $O(2^k)$ input regions with only $O(N)$ parameters. Distributed representations can be clustered to extract relevant features where component extraction algorithms can find the optimal representation (a dictionary). We use a k-SVD approximation algorithm \cite{k-SVD} to mitigate computational expenses.

We consider four physicochemical structure and property prediction tasks from a well-established protein benchmark from \cite{rao2019evaluating} (represented in Figure 1).  Secondary structure prediction involves predicting secondary structure $y \in \{\text{Helix, Strand, Other}\}$ for each amino acid $x$ in a given protein sequence. Solubility prediction considers mapping an  input protein sequence $x$ to a label of $y \in \{\text{Membrane-Bound, Water Soluble}\}$. Homology detection is a sequence classification task, where each input protein $x$ is mapped to a label $y \in \{{1, . . . , 1195}\}$, representing different possible protein folds. Stability prediction is a regression task. We further consider three biomedically relevant function prediction tasks, which are sequence classification tasks (represented in Figure 1). Using R2DL, we predict for a given sequence $x$, its binary class label $y \in \{ \text{AMP, non-AMP} \}$ for antimicrobial-nature prediction \cite{das2021accelerated} or $y \in \{ \text{Toxic, non-Toxic}\}$ for toxicity prediction \cite{das2021accelerated}. Finally, we predict antigen and non-specific binding of antibody variant sequences from \cite{makowski2022co}: given a sequence $x$, the task is to predict $y \in \{\text{on-target, off-target}\}$. Further details on the protein tasks and datasets are in the Method section. The sizes of the individual datasets vary between 4,000 and 50,000 (see supplementary for details on data sizes and train-test splits). Data efficiency is defined as the ratio of the R2DL prediction accuracy to the number of biological sequences used during pretraining and finetuning. We use data efficiency as a metric to compare the performance of R2DL to established benchmarks for the protein tasks in \cite{rao2019evaluating, das2021accelerated, makowski2022co}. For classification tasks, we evaluate prediction accuracy with a top-n accuracy, where $n$ is the number of classes in the protein sequence classification task. For regression tasks, we evaluate prediction accuracy with Spearman's correlation.  

\subsection*{Model Baselines and Data}

%\RV{data size, model size, pretraining corpus size, evaluation metric, task, supervised data size, type of baseline, supervised/unsupervised training approaches}

The baseline models we consider in this work are of two types. Firstly, we consider models trained in a supervised manner, by training standard sequence Long Range Short Term Memory (LSTM) models from scratch. For each downstream peptide or protein classification task, we have labeled (supervised) datasets. The results of these models are reported in Figure 3(a). Secondly, we consider models that are pretrained in an unsupervised manner on protein sequence data and are fintuned for a particular downstream task. Pretraining methods that do not use labeled data pose an advantage, as those models can then learn from a significantly larger number of data samples. In the cases of the toxicity and antimicrobial prediction tasks, the baseline model we compare to has been pretrained on a subset of the UniProt database where sequences are limited to being 50 residues long \cite{uniprot2019uniprot}. The pretraining corpus size is then 1.7 million peptide sequences. Using unlabeled data for pretraining is thus much more advantage than pretraining in a supervised scheme. Of these 1.7 million sequences, only 9,000 are labeled (0.005\% of sequences). The model is a Wasserstein Autoencoder, which is a generative model that undergoes unsupervised pretraining on the subset of UniProt data. The WAE embeddings of the labeled sequences are then  used to train a logistic regressor model on the labeled dataset to obtain a binary classifier for Antimicrobial/non-Antimicrobial (6489 labeled samples) or for toxic/non-toxic (8153 labeled samples) label prediction. For the physicochemical property prediction tasks, the baseline model we consider is pretrained on the Pfam corpus \cite{el2019pfam}. This corpus consists of 31 million protein domains and is widely used in bioinformatics pipelines. Sequences are grouped by protein families which are categorized by evolutionarily-related sequences. In contrast, the downstream physicochemical tasks of structure, homology, stability and solubility prediction have labeled datasets that range from 5,000 to 50,000 samples which the model can be finetuned on. Pretraining thus poses the advantage of modeling the density over a range of protein families and structures, but stipulates that there must be sequence datasets that contain structural and functional information about the downstream task datasets, and typically be of a size on the order of millions of sequences. R2DL eliminates this requirement by repurposing existing pretrained English language models, and leveraging transferrable information from models that are not conditioned on protein sequence information.

\subsection*{Data Efficiency and Accuracy of Reprogramming} % Compared to Pretraining}
%\begin{figure}[h!]
%\centering
%\includegraphics[scale=0.8]{figures/data_eff.png}
%\caption{(a) Training data size and test accuracy of different models (b) Data efficiency of R2DL vs. %pretraining.}
%\end{figure}

\begin{figure}[h!]
     \centering
     \begin{subfigure}[b]{.55\textwidth}
         \centering
         \includegraphics[width=\linewidth]{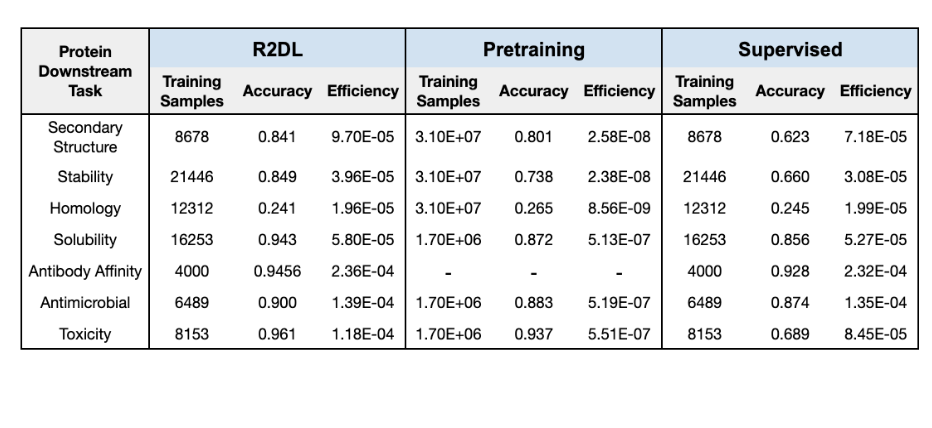}
         \caption{Downstream supervised protein task dataset sizes and test accuracy of the 3 comparable methods introduced in Figure 1.}
         \label{fig:y equals x}
     \end{subfigure}
    \hspace{1mm}
     \begin{subfigure}[b]{.43\textwidth}
         \centering
         \includegraphics[width=\linewidth]{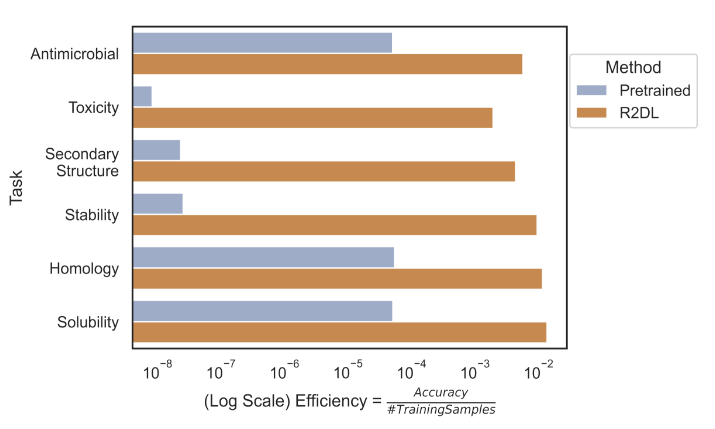}
         \caption{Data efficiency of R2DL vs. pretrained methods as illustrated in Figure 1.}
         \label{Results}
     \end{subfigure}
     \hfill
     \begin{subfigure}[b]{.47\textwidth}
         %\centering
         \hspace{-2mm}
         \includegraphics[scale=0.37]{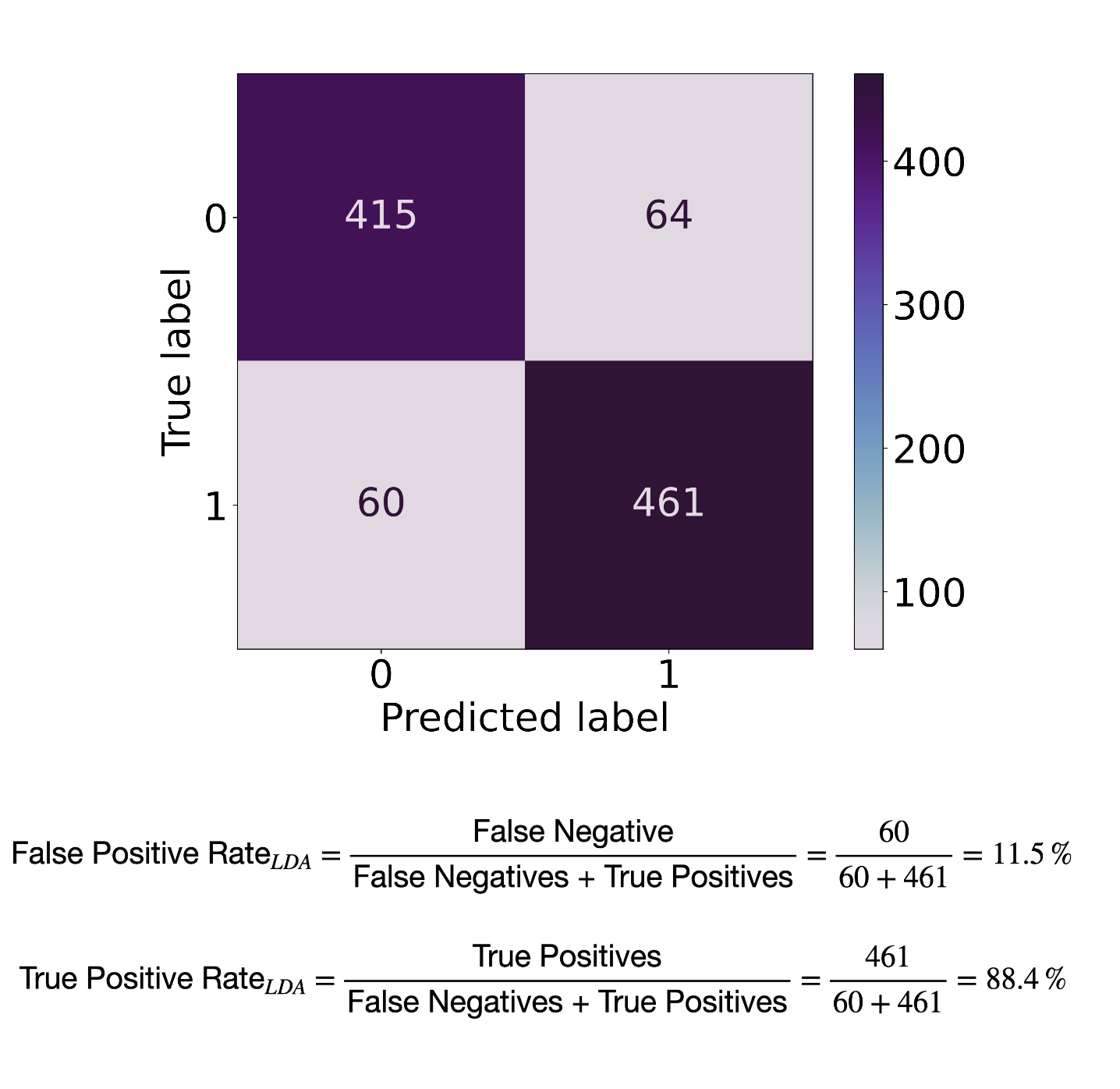}
         \caption{Confusion matrix of the baseline model trained in \cite{makowski2022co} for the antibody affinity prediction task.}
         \label{Results}
     \end{subfigure}
        \hspace{5mm}
     \begin{subfigure}[b]{.47\textwidth}
         %\centering
         \hspace{-2mm}
         \includegraphics[scale=0.37]{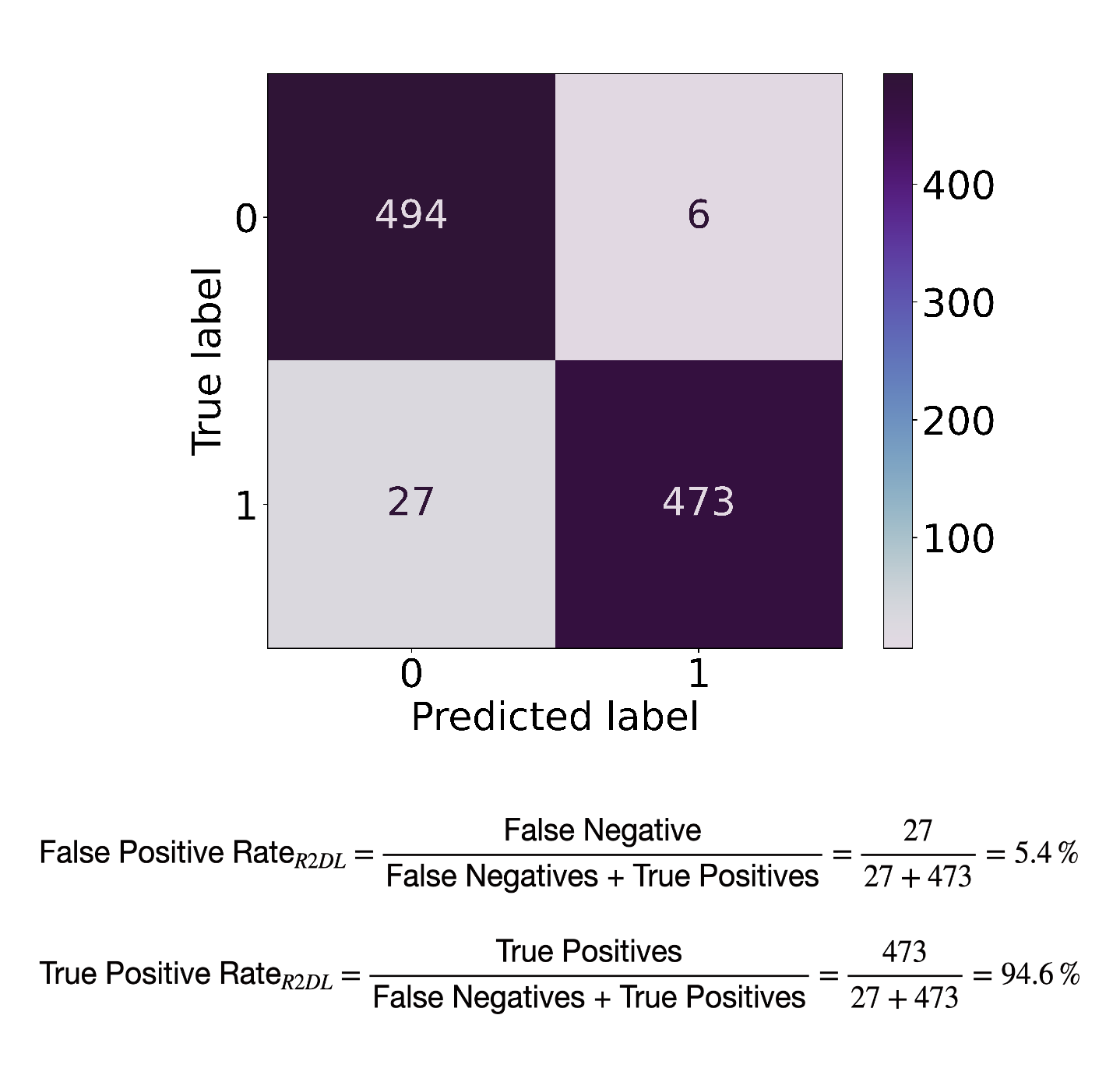}
         \caption{Confusion matrix of the R2DL model for the antibody affinity prediction task.}
         \label{Results}
     \end{subfigure}
     
% \RV{Fix color gradient scales for (c) and (d)}
    \caption{
    Task-specific evaluation of R2DL performance compared to the performance of the baseline models. In Figure 3(a), results for the pretrained baseline models are from unsupervised pretrained transformers for secondary structure, stability, homology, and solubility prediction tasks \cite{rao2019evaluating}. The  baseline models for the antimicrobial and toxicity prediction tasks are logistic regressors trained using sequence embeddings from the pretrained peptide wassertein variational autoencoder \cite{das2021accelerated}. Results for the supervised classifiers are from sequence-level LSTMs trained from scratch on the downstream protein prediction data. For classification tasks, we evaluate prediction accuracy with a top-n accuracy, where $n$ is the number of classes in the protein sequence classification task. For regression tasks, we evaluate prediction accuracy with Spearman's correlation coefficient. Results of the pretrained models on the antibody task dataset have not been previously reported in any work and are hence left out for future work. In 3(b), Data efficiency is defined as the ratio of the R2DL prediction accuracy to the number of protein sequences used during training. In Figure 3(c)-(d), we show a comparison between the performance of a linear discriminant analysis (LDA) model in \cite{makowski2022co} and R2DL on the antibody affinity dataset. The LDA model is a binary classifier which finds the optimal classification boundary by projecting the data onto a one-dimensional feature space and finding a threshold. The antibody affinity dataset consists of 4,000 labeled protein sequences, with labels \{1 (on-target binding), 0 (off-target binding)\}. R2DL achieves a predictive accuracy of 95.5\% compared to the LDA model performance of 92.8\%.}
        \label{fig:}
\end{figure}

% need to add binding task here!
% need one more figure on variance or something?
% report improvement as an avergae improvement over all tasks?
We report the performance of R2DL for the set of 7 protein predictive and their corresponding baselines in Figure 3. Baselines for the  physicochemical prediction tasks are established by a transformer from \cite{rao2019evaluating} that has been pretrained in an unsupervised setting on the Pfam pretraining corpus \cite{punta2012pfam}. Baselines for the antimicrobial and toxicity prediction tasks are established in \cite{das2021accelerated}, where Das et al. pretrained  a Wasserstein Autoencoder on the peptides  from the UniProt corpus \cite{uniprot2019uniprot} using unsupervised training, and then used the latent encodings from autoencoder to train the property classifiers. Baselines for the antibody affinity task are established in \cite{makowski2022co} where they train a linear discriminant analysis model in a supervised setting. Each physicochemical and biomedical function prediction task then has a relatively small, supervised dataset which we split into training and testing sets to train the R2DL framework and evaluate its performance on the test set. Henceforth, we refer to these baselines as task-specific baselines, whereas the baseline model we compare R2DL to varies with the downstream protein prediction task and the best performing model available (see Supplementary for details on task-specific baselines).

We show that, for every prediction task we achieve a higher test accuracy with R2DL than with the corresponding task-specific baseline model when both models are trained on the full labeled dataset. R2DL shows performance improvement up to 11.2\% when compared to the pretrained models, and up to 29.3\% performance when compared to a standard, supervised LSTM that is trained from scratch on the same dataset. However, R2DL needs a pretrained source model and only a small-sized, labeled protein sequence dataset as the input.  And, therefore the size of R2DL training set is limited to the number of samples in the downstream protein prediction dataset. Pretrained models require a large amount of protein sequence data for pretraining, on the order of $10^6$ samples, in addition to the downstream supervised protein task sequence data that the pretrained model is fine-tuned on. In Figure 3(a), we show the number of training samples and corresponding accuracy metric (see Method section for details) of the R2DL, pretrained, and supervised models. In Figure 3(b), we show the data efficiency, \textit{i.e.}, the ratio of the number of training samples (including the pretraining corpus only of biological sequences for pretrained source models) to the accuracy of the model for R2DL and baseline models. We show that R2DL is a maximum of $10^4$ times more data efficient, as in the case of the toxicity prediction task. This is due to the very large number of pretraining data samples required relative to the downstream protein task dataset. %We therefore show that R2DL demonstrates that a sparse encoding is distributed and highly expressive \PD{revise}: we can represent $O(2^k)$ input regions with only $O(N)$ parameters. \PYB{I have no idea how we get this conclusion; let's not go there; Just say R2DL is more efficient} 

Figures 3(c) and 3(d) show the R2DL performance on the antigen affinity prediction task for antibody variant sequences and its comparison with the baseline LDA model reported in \cite{makowski2022co}. R2DL achieves a higher predictive accuracy than the baseline LDA model by 3\% and with a higher classification accuracy with imbalanced datasets. The antibody affinity task dataset has the following distribution {on target: 1516, off-target: 2484}. For 37\% to 62\% class-imbalance ratio of labels, we show that the R2DL model has a better classification accuracy than the LDA model. The learned representations can therefore be inferred to be more accurate in our model than in the baseline model. This is  important, as in many real-world prediction tasks, the dataset is found to be class-imbalanced.

% Distributed representations can be clustered to extract relevant features where component extraction algorithms can find the optimal representation (a dictionary).

\subsection*{R2DL Performance vs. Pretraining Performance in Low Data Settings}

\begin{figure}[h!]
     \centering
     \begin{subfigure}[b]{.3\textwidth}
         \centering
         \includegraphics[width=\linewidth]{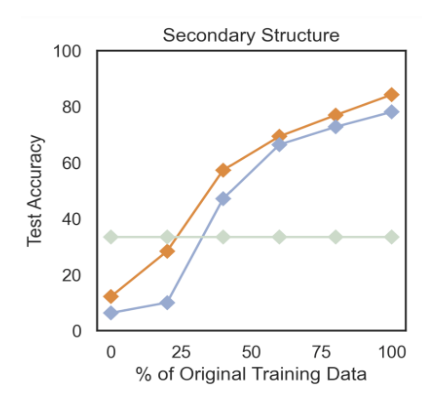}
         \caption{Secondary structure prediction.}
         \label{fig:y equals x}
     \end{subfigure}
    \hspace{5mm}
     \begin{subfigure}[b]{.3\textwidth}
         \centering
         \includegraphics[width=\linewidth]{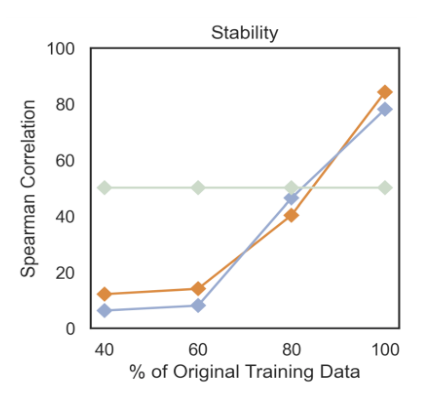}
         \caption{Mutational stability prediction.}
         \label{Results}
     \end{subfigure}
     \hfill
      \begin{subfigure}[b]{.3\textwidth}
         \centering
         \includegraphics[width=\linewidth]{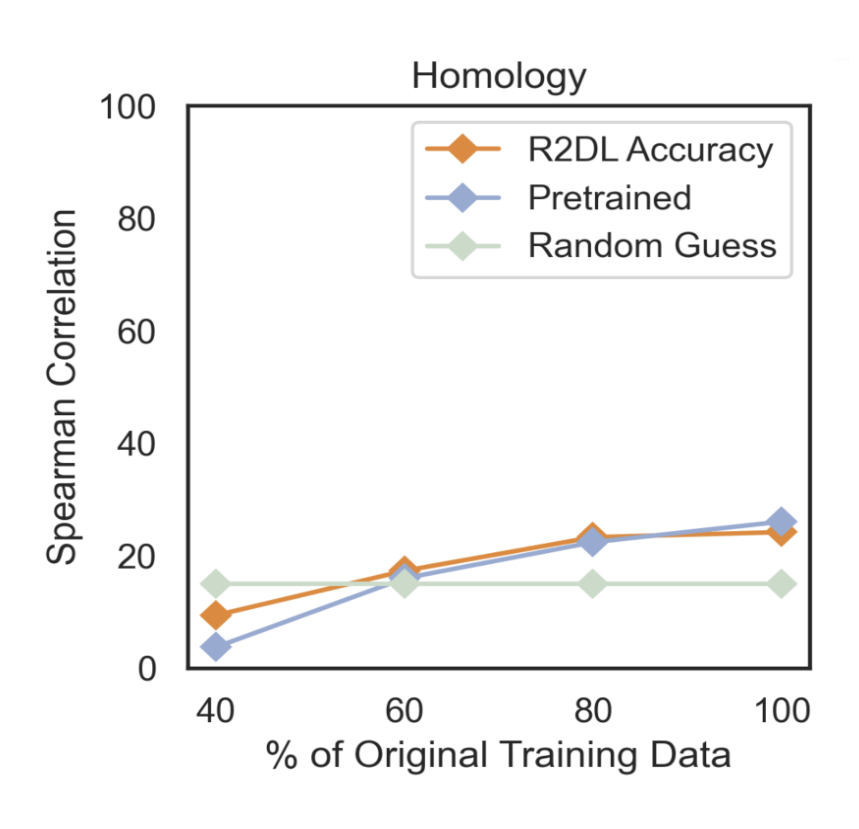}
         \caption{Remote homology prediction.}
         \label{Results}
     \end{subfigure}
     \hfill
     \begin{subfigure}[b]{.3\textwidth}
         \centering
         \includegraphics[width=\linewidth]{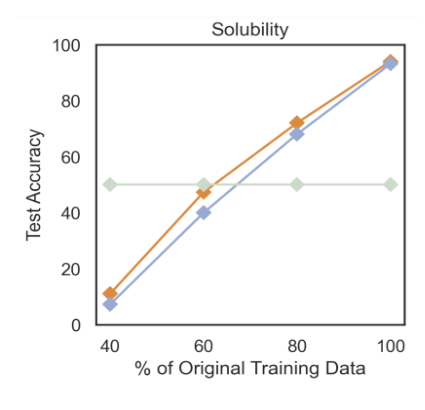}
         \caption{Membrane solubility prediction.}
         \label{Results}
     \end{subfigure}
        \hspace{5mm}
     \begin{subfigure}[b]{.3\textwidth}
         \centering
         \includegraphics[width=\linewidth]{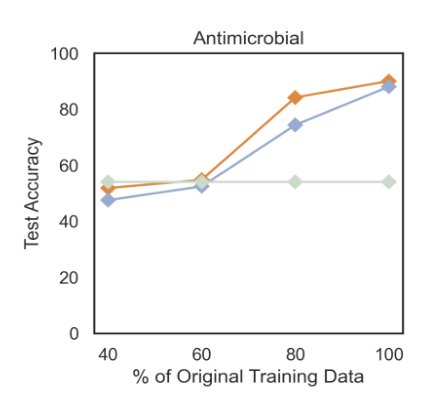}
         \caption{Antimicrobial-nature prediction.}
         \label{Results}
     \end{subfigure}
     \hfill
     \begin{subfigure}[b]{.3\textwidth}
         \centering
         \includegraphics[width=\linewidth]{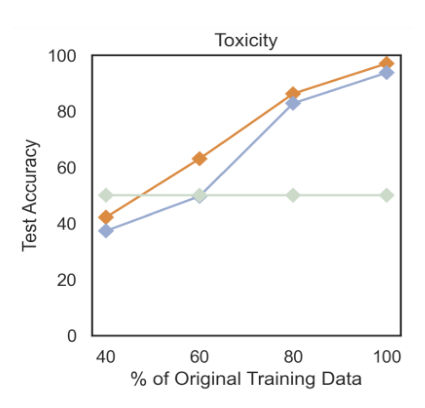}
         \caption{Toxicity prediction.}
         \label{Results}
     \end{subfigure}
     \caption{Results of the R2DL model and baseline model for each  downstream task in reduced training data settings.}
\end{figure}  

Motivated by the data efficiency of R2DL as a framework, we tested the task-specific predictive performance  of R2DL in reduced-data training settings. We compared these results to the performance of task-specific baseline models, when trained and tested in the same restricted data setting. In Figure 4, we show the performance of the R2DL model and then baseline model when trained on 100\%, 80\%, 60\%, and 40\% of a specific task dataset. We show results for the Antimicrobial, Toxicity, Secondary Structure, Stability, Homology, and Solubility prediction tasks in Figure 4 and compare the performance of R2DL and pretrained models against the performance of a random guess. We observe, that for downstream tasks of Toxicity, Secondary Structure, Homology and Solubility, R2DL always performs better than a pretrained protein language model across the size range of  the limited datasets. Furthermore, we observe that, except in the stability task, the rate of failure to perform better than a random guess is higher for the pretrained models than for R2DL. In both cases, R2DL outperforms pretraining until the cutoff point that is the intersection of the random guess curve with the accuracy curves (the point at which the model is not learning any meaningful representation).

\subsection*{Correlation Between Learned Embeddings and Evolutionary Distances}

Beyond comparing the R2DL model against the individual protein task benchmarks, we demonstrate that the R2DL dictionary learning framework shows interpretable correspondences between the learned embeddings in the latent space and the specific protein property. We show this result for the antibody affinity, secondary structure, and toxicity prediction tasks. Figures 5(a-c) show the t-SNE projection of task-specific R2DL embeddings $V_T = \Theta V_S$ of protein sequences for secondary structure, toxicity, and antibody affinity prediction tasks. Clear separation between different protein classes is evident. 
%\PYB{I think we are showing the sentence embedding, not token embedding; no need to mention $V_T$ and $V_S$ here}
We further calculate the similarity between the euclidean distance between the latent representation at the last layer for each amino acid embedding, and compare it to the pairwise evolutionary distance with the BioPython module. In Figure 5(d), we show the euclidean distances between the latent embeddings learned in the R2DL model and the pairwise evolutionary distances between protein sequences, as estimated using BLOSUM62 matrix implemented in the pairwise function of BioPython modulde.%We infer \PD{based on what?} that R2DL captures the evolutionary similarity between comparing empirical sequences and latent vectors to evaluate the quality of the representations.

%\subsubsection*{Antibody Affinity Prediction}

%\subsubsection*{Secondary Structure and Toxicity Prediction}
%\begin{figure}[h!]
%\centering
%\includegraphics[scale=0.85]{figures/biomed_tsne.png}
%\caption{}
%\end{figure}

\begin{figure}[h!]
     %\centering
     \hspace{-2mm}
     \begin{subfigure}[b]{.45\textwidth}
         \centering
         \includegraphics[width=\linewidth]{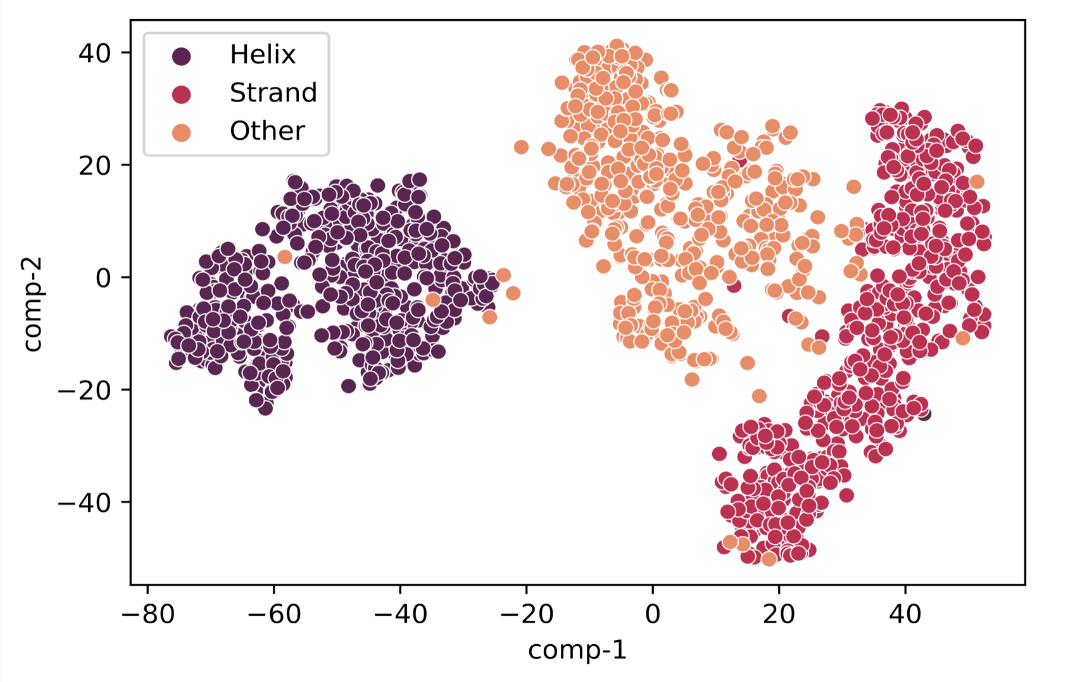}
         \caption{t-SNE clustering plot for secondary structure prediction.}
         \label{fig:y equals x}
     \end{subfigure}
    \hspace{14mm}
     \begin{subfigure}[b]{.45\textwidth}
         \centering
         \includegraphics[width=\linewidth]{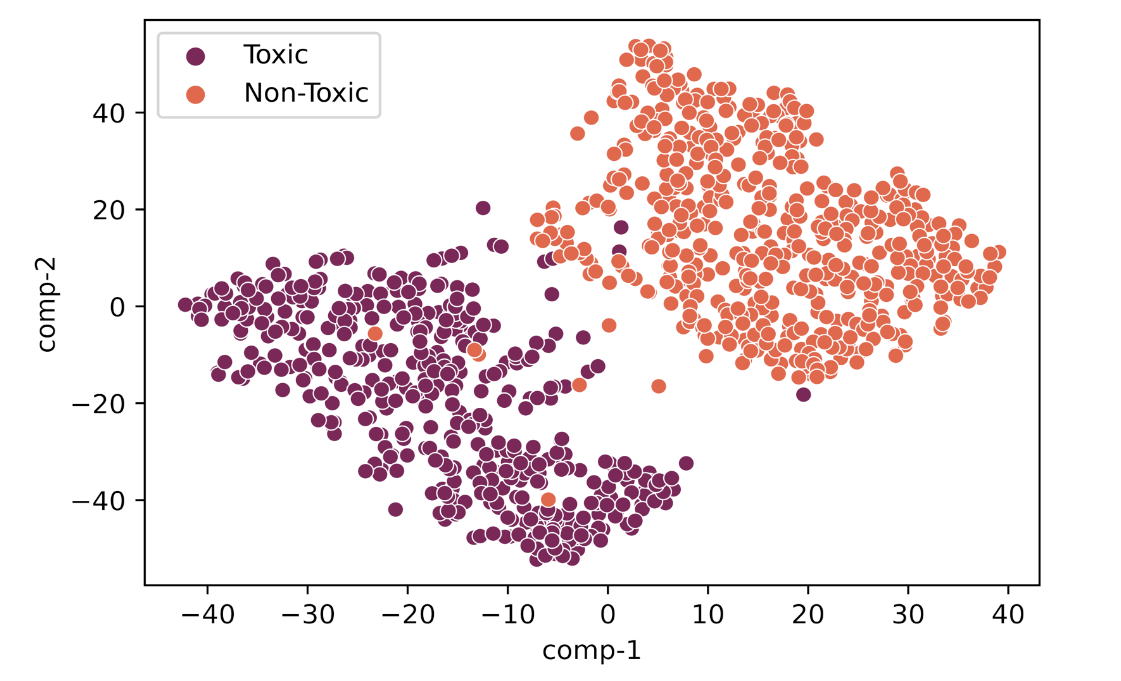}
         \caption{t-SNE clustering plot for toxicity prediction. \newline}
         \label{Results}
     \end{subfigure}
        %\hspace{-15mm}
        \hfill
     \begin{subfigure}[b]{.45\textwidth}
         \centering
         \includegraphics[width=\linewidth]{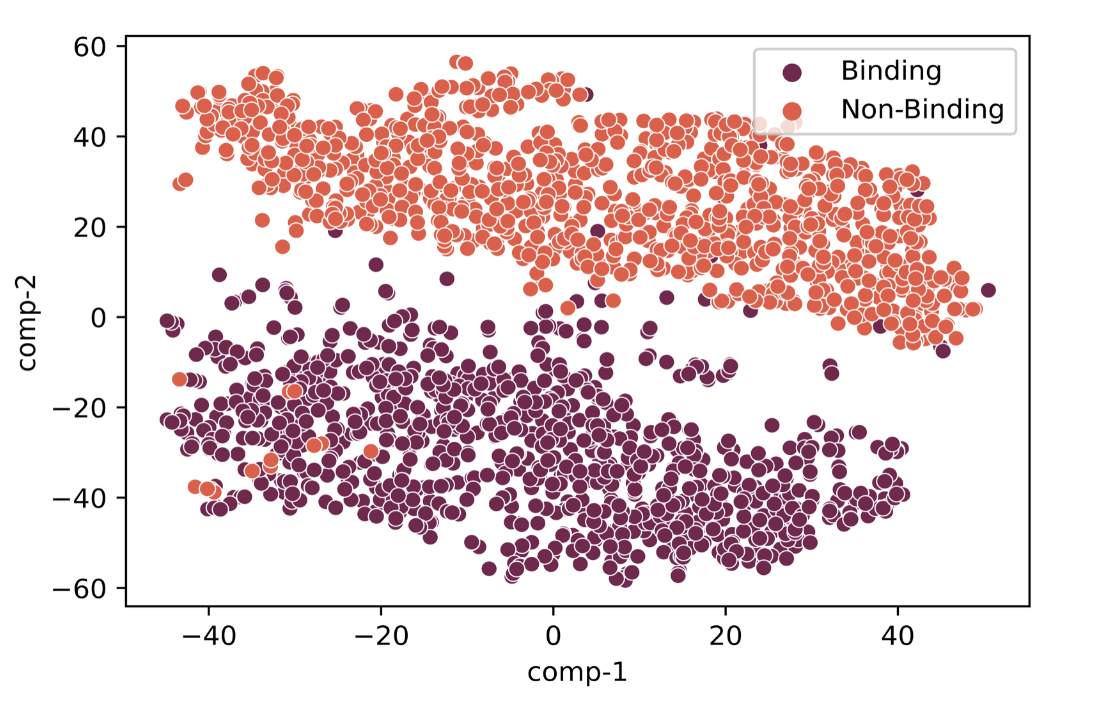}
         \caption{t-SNE plot for antibody affinity prediction. \newline \newline}
         \label{Results}
     \end{subfigure}
        \hspace{13mm}
        %\hfill
     \begin{subfigure}[b]{.5\textwidth}
         \centering
         \includegraphics[width=\linewidth]{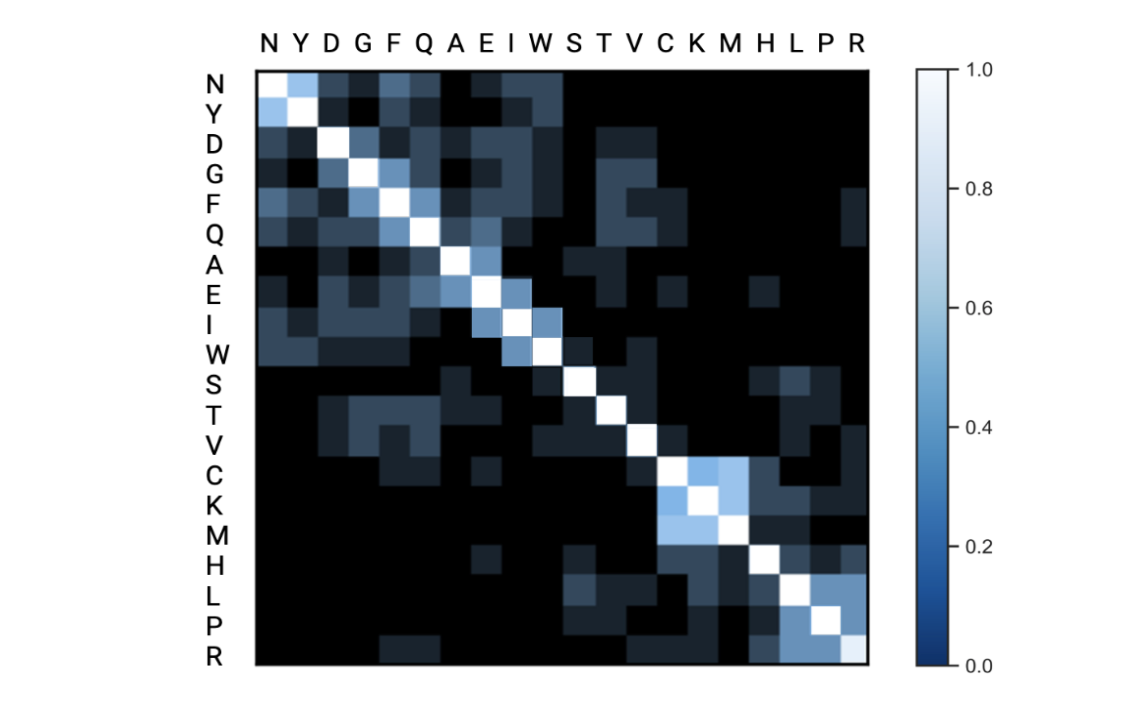}
         \caption{Correlation plot for pairwise evolutionary distances vs. pairwise euclidean distances in R2DL embeddinng space for antibody affinity prediction.}
         \label{Results}
     \end{subfigure}

    \caption{(a-c) Clustering of R2DL learned embeddings for secondary structure prediction, toxicity prediction, and antibody affinity prediction tasks. When tagged by protein property classification, we see very high correspondence between the clusters and protein sequences with the same physicochemical or biomedical property classification. (d) For the antibody affinity prediction task, we observe a high correlation coefficient along the diagonal. This shows that the representation learned by R2DL is highly similar to empirical observations of pairwise residue correlations.}

        \label{fig:}
\end{figure}

%In Figure 6(a-c), we show t-SNE projection of task-specific R2DL  embeddings $V_T = \Theta V_S$ of protein sequences. Clear separation between different protein classes is evident. %when projected with dimensionality reduction such that we can see the structural similarity between the antigen and non-specific binding targets, that 
%the t-SNE clustering for secondary structure prediction and toxicity prediction. Both plots are constructed by taking the learned representation, $V_T = \Theta V_S$ and clustered with dimensionality reduction. 

 % This result also shows the strong correspondence between learned latent features that indicate \PD{what do you mean by structural?}structural similarity between data samples that are reprogrammed to the class-mapped labels. In figure 6\PD{(d)}, we show the correlation matrix between \PD{consider revising}evolutionary pairwise features from the BioPython module and the euclidean distances between learned features in the R2DL framework. 
The matrix shows a correlation of close to 1.0 along the diagonal showing a perfect correspondence between the learned representation and the empirical observations of amino acid relatedness. R2DL thus captures the underlying structure of the linear sequence of amino acid residues in protein sequences in the context of the protein task reprogrammed.

\section*{Discussion}

We propose a new framework, R2DL, to reprogram large language models for various protein tasks. R2DL demonstrates powerful predictive performance across tasks that involve evolutionary understanding, structure prediction, property prediction and protein engineering. We thus provide a strong alternative to pretraining large language models on upto $10^6$ protein sequences. With only a pretrained natural language model (which are abundantly available at the time of writing), a small-sized labeled protein data set of interest, and a small amount of cross-domain finetuning, we can achieve better performance for each protein prediction task with interpretable correspondences between features. Beyond improvements in predictive performance, we show that the ratio of performance improvements to pretraining and training samples involved in the R2DL framework make R2DL up to 105 times more data-efficient than any current methods. This work opens many doors to biological prediction tasks that can acquire very few labeled, high quality data samples. We emphasize  the results of the data-efficiency of R2DL, when applied to biomedically relevant protein predictions, which are critical to advancing scientific understanding and discovery, but have been unsuccessful until now. 

While R2DL does make gradient updates in the framework, the data and resource requirements of the R2DL method is much lower than any unsupervised or self-supervised pretraining approach for protein sequence modeling. Though R2DL has the same data and resource requirements as any standard supervised training approach, R2DL demonstrates much higher task accuracy across a broad and diverse range of property prediction tasks. We claim that R2DL is able to do this because it can leverage the deep representational capacity induced by reprogramming, which standard supervised models cannot achieve without an unjustifiably large number of parameters. R2DL is thus more efficient than existing baseline models in the following aspects: (i) R2DL only requires a pretrained transformer (trained on English language data) and a small-sized, labeled protein sequence data set of interest. We do not make any updates to the pretrained model itself, unlike traditional transfer learning methods. Rather we make updates to the R2DL model during a supervised training process that optimizes over class-mapped labels. (ii) R2DL does not require large-scale  un/self-supervised pretraining on millions of unlabeled protein sequences, as in \cite{rao2019evaluating, das2021accelerated, meier2021language}. (iii) Further, R2DL does not require any large-scale supervised pretraining, which has been found beneficial in protein-specific tasks \cite{rao2019evaluating} as well as in computer vision \cite{dosovitskiy2020image}. Labeling protein sequences at scale, particularly for biomedical function, is almost infeasible for the size of dataset that is required for supervised pretraining. With these three considerations in mind, we pose R2DL as a data-efficient alternative to pretraining methods for  protein prediction tasks of biological and biomedical relevance. To the best of our knowledge,  R2DL is the first framework without explicit pretraining that facilitates accurate predictions across a general suite of protein prediction tasks and provides interpretable correspondences between amino acid features that are very closely aligned with domain knowledge (evolutionary distances). The success of R2DL can be attributed to its representational power to encode a sparse representation by leveraging the natural language modeling entailed in large language models for efficient learning on protein structure and function prediction tasks, as both English and protein sequences follow Zipf’s law \cite{newman2005power}.\\

We first demonstrate the effectiveness of R2DL on a set of physicochemical structure and property prediction tasks, and then on a set of biomedically relevant function prediction tasks, for  protein sequences. We show predictive performance improvements against pretrained methods (up to 11\% in stability) and standard supervised methods (up to 3.2\% in antibody affinity). Similarly, on the remaining tasks, we show performance improvements over the best reported baseline in structure prediction (4.1\%), homology (2.3\%), solubility (7.1\%), antibody affinity (3.2\%), toxicity (2.4\%). %The accuracy results of individual protein tasks are reported in Table 3(a). In addition to the accuracy, we show the ratio of accuracy to the number of pretraining samples used in the state-of-the-art benchmarked models for each individual protein task which are reported in Figure 3(b). 
R2DL thus shows the capability to learn a general representation of protein sequences that can be efficiently adopted to different downstream protein tasks. These powerful representation capabilities as evidenced by its ability to achieve high performance across protein datasets with a highly varied number of task-specific training samples. The performance of R2DL across protein tasks show the potential to repurpose and develop powerful models that can learn from small, curated, and function-specific datasets. This mitigates the need to train large pretrained models for peptide learning tasks. We thus provide an alternative method to pretraining that is cheaper to run and more accurate, and therefore adoptable to broader researcher communities who may not have access to large-scale compute. This potential is critical for many applications, such as discovery of new materials, catalysts, as well as  drugs. Although we establish the efficacy and efficiency of R2DL in a domain where pretrained large language models already do exist, we hope that our work will pave the path to extending this approach to other domains where pretrained LLMs do not exist, such as polymers.

\section*{Method}

\subsection*{Representation of Tokens}

In the R2DL framework, we use 2 input datasets, an English language text dataset (source dataset) and a protein sequence dataset (target dataset). The vocabulary size of a protein sequence dataset at a unigram level is 20, as proteins are composed of 20 different natural  amino acids. We obtain a latent representation of the English text vocabulary, $V_S$, by extracting the learned embeddings of the data from a pretrained language model (source model). The protein sequence data is embedded in the same latent space, and is termed the target vocabulary, $V_T$. For each task, the token embedding matrix is of dimensions $(n, m)$ where $n$ is the number of tokens and $m$ is the length of the embedding vectors.
%the maximum length of amino acid sequences with each English-alphabetical character corresponding to 1 of the 20 protein-building amino acids. 
We use the same encoding scheme of $V_S$ and $V_T$ across all downstream tasks. 

\subsection*{Procedure Description of the R2DL Framework for a Protein Task}

\begin{itemize}
    \item \textbf{Procedure Inputs}: Pretrained English sentence classifier $\mathbf{C}$, target model training data $\mathbf{X_\ell}$ for task $\ell$, class mapping label function, $h_\ell$ (if classification) where  \\ $\ell \in \{\text{Secondary Structure, Fluorescence,
    Homology, Solubility, Antimicrobial, Toxicity, Antibody}\}$.
    
    \item \textbf{Procedure Hyperparameters}: Maximum number of iterations $T_1$ for updates to $\Theta$, number of iterations $T_2$ for k-SVD, step size $\{\alpha_t\}^{T_1}_{t=1}$
    
    \item \textbf{Procedure Initialization}: Random initialization of $\Theta$, obtain the source token embedding matrix $V_S$
    
    \item \textbf{Define Objective Function}: Objective function for k-SVD: $\| V_T - \Theta V_S \| \leq \epsilon$

    \item \textbf{k-SVD Approximation of $\Theta$}: If $ t_1 \leq T_1$, while {$t_2 \leq T_2$}  use approximate k-SVD to solve $V_T \approx  \Theta V_S$, \; $t_2 \longleftarrow t_2 + 1$ \;

    \item \textbf{Calculate the Loss and Perform Gradient Descent}: $\Theta \longleftarrow \Theta - \alpha_{t} \cdot \nabla_\Theta \text{Loss}(\Theta, \mathbf{X}_\ell, h_\ell, \mathbf{C}) $\;, $t_1 \longleftarrow t_1 + 1$ \ and return to the previous K-SVD step

    \item \textbf{Output Protein Sequence Labels for Protein Sequence $x$ of Task $\ell$}: $h_\ell(\mathbf{C}(\Theta,x))$

\end{itemize}

%To reprogram the pretrained classifier, we use an embedding mapping $f_\theta : s_i \longmapsto t_i$ where $s_i \in V_S$ and $t_i \in V_T$. Dimension of the input space of $V_S$ and $V_T$ is $|V_S|$, and $|V_T|$ respectively, where $|V_T| \ll |V_S|$. The mapping function is parametrized by $\Theta \in \mathbf{R}^{a \times b}$, which represents the coefficients of the atoms in $V_S$ such that $V_T = \theta V_S$. The observation of $|V_T| \ll |V_S|$ requires that our mapping has high representational capacity, so we encode a sparse representation of $V_T$, to extract relevant features from the source vocabulary embeddings $\{V_S\}^{|V_S|}_{a=1}$ for the alternate task. To that end, approximate the dictionary, we use a k-SVD solver to optimize over the cross entropy loss for updates to $\theta$. (this is repeated in R2DL Training and Optimization Procedure)

We are given a pretrained English classifier, $\mathbf{C}$, and a protein sequence target-task dataset $\mathbf{X}_\ell$. We denote the task with $\ell$, such that $\ell \in \{\text{Secondary Structure, Fluorescence, Homology, Solubility, Antimicrobial, Toxicity, Antibody}\}$. We also encode an output label mapping function $h_\ell$ specifying the one-to-one correspondence between source and target labels. 
As shown in Figure 2, the source vocabulary embedding, $V_S$, is extracted from the pretrained model, $\mathbf{C}$. The next objective is to learn $\Theta$ that approximates the embedding of tokens in $\mathbf{X}_\ell$ (denoted by $V_T$) in the representation space of the source model.

We aim to learn $\Theta \in \mathbf{R}^{a \times b}$ that finds the optimal coefficients $\{\theta_t\}$ for each of the target tokens $t \in \{1,..., a\}$ in $V_T \in \mathbf{R}^{a \times m}$ to be represented as a sparse encoding of the dictionary, $V_S \in \mathbf{R}^{b \times m}$, such that $V_T = \Theta V_S $. For a given target protein sequence $x$ from the $\ell$-th task,
$\Theta$ is used to perform the target task through the transformation $h_\ell(\mathbf{C}(\Theta,x))$. While we do not make any modification to the parameters or architecture of $\mathbf{C}$, we assume access to the gradient $\nabla_{\Theta} \text{loss($\cdot$)}$ for loss evaluation and parameter updates during training.

A target token embedding $v_t \in \mathbf{R}^{m}$ can be represented as a sparse linear combination of the source token embeddings (rows) in $V_S$, $v_t = \theta_t V_s $. $v_t$ is the representation of the protein token in the dictionary space and satisfies $||v_t - \theta_t V_s||_p \leq \epsilon$, where $\|\cdot\|_p$ is an $L_p$ norm and $\theta_t$ is made to be sparse by satisfying $||\theta_t||_0 \leq k$ for all $t$. An exact solution $v_t = \theta_t V_S$ is computationally expensive to find, and is subject to various convergence traps, so for the purpose of our efficient fine-tuning approach we approximate $v_t \approx \theta_t V_S $ using k-SVD. We first fix the dictionary $V_S$, as extracted from $\mathbf{C}$, and then find the optimal $\Theta$ according to the optimization problem, by minimizing the alternative objective $\sum_{t=1}^a ||\theta_t||_0$
subject to
$\|V_T - \Theta V_S\|^{2}_{F} \leq \epsilon$ 
as explored in \cite{k-SVD}. While algorithms exist to choose an optimal dictionary (an exact solution to k-SVD) that can be continually updated \cite{k-SVD}, we penalize computational expense over performance for the purpose of maintaining an efficient solution (at the cost of statistically insignificant improvements in accuracy) by using a predetermined number of iterations for k-SVD convergence, which is then used to evaluate the cross entropy loss on $h_\ell(\mathbf{C}(\Theta,x))$ and update the mapping function $\Theta$.

\subsection*{Data}

\subsubsection*{Classification}

We provide five biologically relevant downstream physicochemical property prediction tasks, adapted from \cite{rao2019evaluating} to serve as benchmarks. We categorize these into property prediction, structure prediction, evolutionary understanding, and protein engineering tasks. The sizes of the individual datasets vary between 4,000 and 50,00 (see supplementary for details on data sizes and train-test splits).
\\

\textbf{Secondary Structure Prediction (Structure Task):} Secondary structure (SS) is critical to understanding the function and stability of a protein, and SS prediction is an important intermediate step in designing designing protein complexes. This dataset, obtained from \cite{klausen2019netsurfp} has 8,678 data samples. It is derived from the CB513 dataset, and each amino acid, $x$ in a protein sequence is mapped to $y \in \{\text{Helix, Strand, Other}\}$. The benchmark for this task is a transformer that reports a best performance of 80\% accuracy. \\

\textbf{Solubility:} This task takes an input protein $x$ and maps it to a label of $y \in \{\text{Membrane-Bound, Water Soluble}\}$. Determining the solubility of proteins is useful when designing proteins or evaluating their function for particular cellular tasks. This dataset, obtained from \cite{almagro2017deeploc} has 16,253 data samples. The benchmark is a pretrained transformer, that achieves a best performance of 91\% on a binary classification task. \\

\textbf{Antigen Affinity (Protein Engineering):} Therapeutic antibody development requires the selection and engineering of molecules with high affinity and other drug-like biophysical properties. This dataset, obtained from \cite{makowski2022co} has 4,000 data samples. The task is to map an input protein $x$ to a label $y \in \{\text{on-target, off-target} \}.$ The task corresponds to predicting antigen and non-specific binding. The benchmark for this task is a Linear Discriminant Analysis model with Spearman's $\rho$ values for antigen binding (0.87) and for non-specific binding (0.67). \\

\textbf{Antimicrobial Prediction (AMP) (Property Task):} Determining the antimicrobial nature of a peptide is a critical step in developing antimicrobials to fight against resistant pathogens. The dataset, obtained from \cite{das2021accelerated}, consists of 6,489 labeled protein sequences $x$, is mapped to a label $y \in \{\text{AMP, non-AMP}\}$. The original model trained on this data provides a de novo approach for discovering new, broad-spectrum and low-toxic antimicrobials. The benchmark for this task is a transformer that reports a best performance of 88\% accuracy with a pretrained classifier. \\

\textbf{Toxicity (Property Task):} Improving the functional profile of molecules, especially in the context of drug discovery, requires optimizing for toxicity and other  physicochemical properties. To that end, toxicity is an important property to predict in AMP development. This dataset, obtained from \cite{das2021accelerated} consists of 8,153 antimicrobial peptide sequences which are either toxic (positive class), or non-toxic (negative class). The benchmark for this task is a transformer that reports a best performance of 93.78\% accuracy with a pretrained classifier. \\

\subsubsection*{Regression}

\textbf{Stability (Protein Engineering Task):} This regression task where each protein, $x_i$ is mapped to $y_i \in \mathbb{R}$ based on maintaining its fold beyond a threshold of concentration. This dataset, obtained from \cite{rocklin2017global} has 21,446 data samples. Stability is an important protein engineering task, as we can use this fold concentration to test protein inputs such that design candidates are stable in the settings of different tasks. The benchmark for this task is a transformer that reports a best performance of 0.73 Spearman's $\rho$. \\

\textbf{Homology (Evolutionary Understanding Task):} This is a sequence classification task where each input protein, $x$ is mapped to a protein fold represented by $y \in \{{1, ..., 1195}\}$. This dataset, obtained from \cite{hou2018deepsf} has 12,312 data samples. Detecting homologs is particularly important in a biomedical context as they inform structural similarity across a set of sequences, and can indicate emerging resistance of antibiotic genes [cite]. The original model removes entire homologous groups during model training, thereby enforcing that models generalize well when large evolutionary gaps are introduced. The benchmark for this task is a LSTM that reports a best performance of 26\% Top-1 Accuracy. \\

\subsection*{R2DL Settings and Hyperparameter Details}

\subsection*{AMP}
 The full AMP dataset size is 8112, we use a training set size of 6489 and a test set size of 812. We use the $L_0$ norm in our objective function, 10,000 k-SVD iterations and $\epsilon = 0.045$.
\subsection*{Toxicity}
 The full Toxicity dataset size is 10,192, we use a training set size of 8153 and a test set size of 1020. We use the $L_0$ norm in our objective function, 10,000 k-SVD iterations and $\epsilon = 0.045$.

\subsection*{Secondary Structure}
 The full Toxicity dataset size is 9270, we use a training set size of 7416 and a test set size of 1854. We use the $L_0$ norm in our objective function, 9,000 k-SVD iterations and $\epsilon = 0.38$.

\subsection*{Stability}
 The full Stability dataset size is 56,126, we use a training set size of 44,900 and a test set size of 11,226. We use the $L_0$ norm in our objective function, 6,000 k-SVD iterations and $\epsilon = 0.29$.

\subsection*{Homology}
 The full Homology dataset size is 13,048, we use a training set size of 10,438 and a test set size of 2,610. We use the $L_0$ norm in our objective function, 4,000 k-SVD iterations and $\epsilon = 0.73$.

\subsection*{Solubility}
 The full Solubility dataset size is 43,876, we use a training set size of 35,100 and a test set size of 8,775. We use the $L_0$ norm in our objective function, 9,000 k-SVD iterations and $\epsilon = 0.42$.

\subsection*{Data and Code Availability}
Links to protein sequence data and code are available on Github (github.com/riavinod/r2dl)

\clearpage

%\appendix{}

\nocite{*}
\bibliography{citations}

\pagebreak

\section*{Supplementary Information}

\begin{table}[!h]
%\setlength\tabcolsep{4pt}
%\begin{tabularx}{\textwidth}{*{6}{Z}}
%\begin{tabular}{p{3cm}p{3cm}p{3cm}p{3cm}p{3cm}p{3cm}p{3cm}}
\begin{tabular}{p{2cm} p{2cm} p{4cm} p{2cm} p{1.5cm} p{4cm}}

\toprule
\multicolumn{6}{c}{}   \\
  {Protein Task} & {Source Model} & {Source Task} & {Regression or Classification} & {Source Labels} & {Target Labels}   \\ \midrule
    Antimicrobial        & Transformer & Sentiment Classification & Classification & Positive, Negative & AMP, non-AMP  \\ 
    Toxicity             & Transformer & Sentiment Classification & Classification & Positive, Negative & Toxic, non-Toxic \\
    Secondary Structure  & Transformer & Sentiment Classification & Classification & Positive, Neutral, Negative & Helix, Strand, Other \\
    Stability            & Transformer & Sentiment Classification & Regression     & -                  & - \\
    Homology             & Transformer & Sentiment Classification & Regression     & -                  & - \\
    Solubility           & Transformer & Named Entity Recognition & Classification & Positive, Negative & Soluble, non-Soluble \\
    Binding              & Transformer & Sentiment Classification  & Classification & Positive, Negative & On-target, Off-target \\
\bottomrule
%\end{tabularx}
\end{tabular}
\caption{Summary of the source and target tasks for reprogramming}

\end{table}

\begin{figure}[h!]
%\centering
\hspace{-8mm}
\includegraphics[scale=0.6]{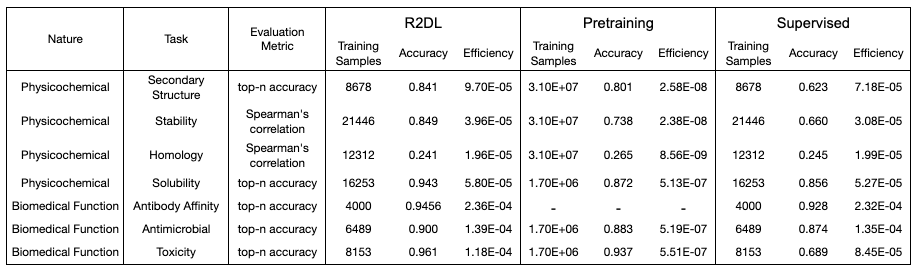}
\caption{Summary of protein prediction tasks and evaluation metrics with model performance.}
\end{figure}

% \SetKwInput{KwInput}{Input}                % Set the Input
% \SetKwInput{KwOutput}{Output}              % set the Output

\subsection*{Model Baselines}

\begin{table}[h!]
\centering
\caption{Toxicity and Antimicrobial-nature reported in \cite{das2021accelerated}.}
\begin{tabular}{@{}lllllll@{}}
\toprule
\multicolumn{1}{c}{\multirow{2}{*}{Attribute}} & \multicolumn{3}{l}{Data-Split} & \multicolumn{2}{l}{Accuracy} &  \\ \cmidrule(lr){2-6}
\multicolumn{1}{c}{}        & Train & Valid & Test & Majority Class & Test & \multicolumn{1}{c}{} \\ \cmidrule(r){1-1} 
\{Toxic, non-Toxic\}        & 8153  & 1019  & 1020 & 0.82 & 0.93 \\
\{AMP, non-AMP\}            & 6489  & 811  & 812 & 0.82  & 0.88  \\ \bottomrule
\end{tabular}
\end{table}

\begin{table}[h!]
\centering
\caption{Structure prediction, Remote Homology, Stability reported in \cite{rao2019evaluating}.}
\begin{tabular}
{>{\centering\arraybackslash}m{5cm}|>{\centering\arraybackslash}m{2cm}|>{\centering\arraybackslash}m{2cm}|>{\centering\arraybackslash}m{2.5cm}>{\centering\arraybackslash}m{2cm}>{\centering\arraybackslash}m{2cm}}
\toprule
Task                             & Model                 & Accuracy Metric         & Test Accuracy  \\ \midrule
Secondary Structure Prediction   & One Hot + Alignment   & Accuracy (3-class)      & 0.80                 \\
Remote Homology Detection        & LSTM                  & Top 1 Accuracy          & 0.26               \\
Stability                        & Transformer           & Spearman's Rho          & 0.73               \\ \bottomrule
\end{tabular}
\end{table}

\begin{table}[h!]
\centering
\caption{Solubility reported in \cite{elnaggar2020prottrans}}
\begin{tabular}
{>{\centering\arraybackslash}m{3cm}|>{\centering\arraybackslash}m{2cm}|>{\centering\arraybackslash}m{2cm}>{\centering\arraybackslash}m{2.5cm}}
\toprule
% type of accuracy measure is impt
Task         & Model                    & Test Accuracy  \\ \midrule
Solubility   & ProtT5-XL-UniRef50	    & 0.91              \\ \bottomrule
\end{tabular}
\end{table}

\begin{table}[h!]
\centering
\caption{Antibody Affinity Binding reported in \cite{makowski2022co}.}
%\label{tab:my-table}
%\begin{tabular}{@{}lllll@{}}
\begin{tabular}
{>{\centering\arraybackslash}m{3cm}|>{\centering\arraybackslash}m{2cm}|>{\centering\arraybackslash}m{2cm}>{\centering\arraybackslash}m{2.5cm}}
\toprule
Task         & Model                    & Test Accuracy  \\ \midrule
Antibody Affinity   & Linear Discriminant Analysis	    & 0.92              \\
 \bottomrule
\end{tabular}
\end{table}

\subsection*{R2DL Results}

\begin{table}[h!]
\centering
\caption{R2DL: AMP Classification}
%\label{tab:my-table}
%\begin{tabular}{@{}lllll@{}}
\begin{tabular}{>{\centering\arraybackslash}m{5.0cm}|>{\centering\arraybackslash}m{1.7cm}|>{\centering\arraybackslash}m{1.7cm}|>{\centering\arraybackslash}m{1.7cm}|>{\centering\arraybackslash}m{1.7cm}}

\toprule
Source Model                     & AMP Sequence Samples & k-SVD Iterations & Training Accuracy & Test Accuracy \\ \midrule
BERT (Bidirectional Transformer) & 6489                 & 100              & 87.12             & 85.64         \\
BERT (Bidirectional Transformer) & 6489                 & 250              & 85.67             & 82.33         \\
Bi-LSTM                          & 6489                 & 100              & 79.40             & 81.90         \\ \bottomrule

\end{tabular}
\end{table}

\begin{table}[h!]
\centering
\caption{R2DL: Toxicity Prediction}
%\label{tab:my-table}
\begin{tabular}{@{}lllll@{}}
%\begin{tabular}{>{\centering\arraybackslash}m{5.0cm}|>{\centering\arraybackslash}m{1.7cm}|>{\centering\arraybackslash}m{1.7cm}|>{\centering\arraybackslash}m{1.7cm}|>{\centering\arraybackslash}m{1.7cm}}

\toprule
Source Model                     & AMP Sequence Samples & k-SVD Iterations  & Test Accuracy \\ \midrule
BERT (Bidirectional Transformer) & 8153                 & 100                           & 87.23         \\
BERT (Bidirectional Transformer) & 8153                 & 250                          & 86.93         \\
Bi-LSTM                          & 8153                 & 100                         & 81.25         \\ 
\bottomrule

\end{tabular}
\end{table}

\begin{table}[h!]
\centering
\caption{R2DL: Secondary Structure Prediction}
%\label{tab:my-table}
\begin{tabular}{@{}lllll@{}}
\toprule
Source Model & Training Samples  & k-SVD Iterations & Training Accuracy & Test Accuracy \\ \midrule
BERT         & 8,678         & 10000                              & 71.47             & 63.65         \\
BERT         & 8,678         & 15000                              & 74.34             & 69.91         \\
BERT         & 8,678         & 20000                               & 76.32             & 74.92         \\ 
\bottomrule
\end{tabular}
\end{table}

\begin{table}[h!]
\centering
\caption{R2DL: Remote Homolgy Detection (Top-1 Accuracy)}
%\label{tab:my-table}
\begin{tabular}{@{}lllll@{}}
\toprule
Source Model & Training Samples & k-SVD Iterations & Training Accuracy & Test Accuracy \\ \midrule
BERT          & 12,312                 & 10000              & 11.34            & 10.76         \\
BERT          & 12,312                 & 15000              & 16.45             & 15.67       \\
BERT          & 12,312                 & 20000              & 26.23             & 24.50         \\ 
\bottomrule
\end{tabular}
\end{table}

\begin{table}[h!]
\centering
\caption{R2DL: Stability (Spearman's Rho)}
%\label{tab:my-table}
\begin{tabular}{@{}lllll@{}}
\toprule
Source Model & Training Samples & k-SVD Iterations & Training Accuracy & Test Accuracy \\ \midrule
BERT          & 53,679                & 10000              & 60.23            & 61.89         \\
BERT          & 53,679                 & 15000              & 68.62             & 67.20       \\
BERT          & 53,679                 & 20000              & 70.78             & 69.73         \\ 
\bottomrule
\end{tabular}
\end{table}

\begin{table}[h!]
\centering
\caption{R2DL: Fluorescence (Spearman's Rho)}
%\label{tab:my-table}
\begin{tabular}{@{}lllll@{}}
\toprule
Source Model & Training Samples & k-SVD Iterations & Training Accuracy & Test Accuracy \\ \midrule
BERT          & 21,446                & 10000             & 61.29            & 52.82         \\
BERT          & 21,446                & 15000             & 61.02             & 59.46       \\
BERT          & 21,446                & 20000             & 70.90             & 62.34         \\ 
\bottomrule
\end{tabular}
\end{table}

\begin{table}[h!]
\centering
\caption{R2DL: Solubility}
%\label{tab:my-table}
\begin{tabular}{@{}lllll@{}}
\toprule
Source Model & Training Samples & k-SVD Iterations & Training Accuracy & Test Accuracy \\ \midrule
TinyBERT          & 6623                & 10000             & 68.93            &   69.82         \\
TinyBERT         & 6623                & 15000             & 87.22             & 89.3       \\
TinyBERT          & 6623                & 20000             & 92.85             & 93.21         \\ 
\bottomrule
\end{tabular}
\end{table}

\subsection*{R2DL Results from the Reduced Training Data Setting}

\subsection{Restricted Training Data Setting}
To further investigate the efficacy of the transfer learning approach, we compare the performance of R2DL versus the model trained from scratch with AMP data, with a restricted training data set. The test accuracy across tasks indicate that R2DL performs better when fewer labeled training data samples are available. Below 25\% of training data samples, both methods approximately do worse than random prediction, so we do not reduce the training data to evaluate performance after this threshold. %\footnote{5000 AMP training samples and below were excluded from the results below as they showed statistically insignificant test accuracy.}.  

\begin{table}[h!]
\centering
\caption{Restricted Data Setting: Toxicity Prediction}
%\label{tab:my-table}
%\begin{tabular}{@{}lllll@{}}
\begin{tabular}{>{\centering\arraybackslash}m{5.0cm}|>{\centering\arraybackslash}m{2.5cm}|>{\centering\arraybackslash}m{2cm}|>{\centering\arraybackslash}m{2.5cm}>{\centering\arraybackslash}m{2cm}}

\toprule
Task            & Training Samples & R2DL Test Accuracy & Bi-LSTM Test Accuracy  \\ \midrule
Toxicity Prediction & 5000                 & 42.12              & 37.34                   \\
Toxicity Prediction & 6000                 & 62.98              & 49.62                     \\
Toxicity Prediction & 7000                 & 86.23              & 82.78                      \\
Toxicity Prediction & 8153                 & 89.34              & 93.7                     \\
 \bottomrule

\end{tabular}
\end{table}

\begin{table}[h!]
\centering
\caption{Restricted Data Setting: AMP Prediction}
%\label{tab:my-table}
%\begin{tabular}{@{}lllll@{}}
\begin{tabular}{>{\centering\arraybackslash}m{5.0cm}|>{\centering\arraybackslash}m{2.5cm}|>{\centering\arraybackslash}m{2cm}|>{\centering\arraybackslash}m{2.5cm}>{\centering\arraybackslash}m{2cm}}

\toprule
Task            & Training Samples & R2DL Test Accuracy & Bi-LSTM Test Accuracy \\ \midrule
AMP Prediction  & 3500                 & 59.82              & 64.52                   \\
AMP Prediction  & 4500                 & 72.76              & 68.41                     \\
AMP Prediction  & 5500                 & 84.17            & 74.34                      \\
AMP Prediction  & 6489                 & 90.01            & 88.0                    \\
 \bottomrule

\end{tabular}
\end{table}

\begin{table}[h!]
\centering
\caption{Restricted Data Setting: Secondary Structure Prediction (SSP)}
%\label{tab:my-table}
%\begin{tabular}{@{}lllll@{}}
\begin{tabular}{>{\centering\arraybackslash}m{5.0cm}|>{\centering\arraybackslash}m{2.5cm}|>{\centering\arraybackslash}m{2cm}|>{\centering\arraybackslash}m{2.5cm}>{\centering\arraybackslash}m{2cm}}

\toprule
Task    & Training Samples     & R2DL Test Accuracy & Bi-LSTM Test Accuracy \\ \midrule
Structure Prediction     & 3378                 & 12.09              & 06.23                   \\
Structure Prediction     & 4478                 & 34.26              & 37.93                     \\
Structure Prediction     & 6678                 & 69.28              & 66.34                      \\
Structure Prediction     & 8678                 & 84.14              & 78.0                    \\
 \bottomrule

\end{tabular}
\end{table}

\begin{table}[h!]
\centering
\caption{Restricted Data Setting: Remote Homology Detection}
%\label{tab:my-table}
%\begin{tabular}{@{}lllll@{}}
\begin{tabular}{>{\centering\arraybackslash}m{5.0cm}|>{\centering\arraybackslash}m{2.5cm}|>{\centering\arraybackslash}m{2cm}|>{\centering\arraybackslash}m{2.5cm}>{\centering\arraybackslash}m{2cm}}

\toprule
Task    & Training Samples     & R2DL Test Accuracy & Bi-LSTM Test Accuracy \\ \midrule
Homology     & 4312                 & 09.35              & 03.69                   \\
Homology     & 8312                 & 17.26              & 15.93                     \\
Homology     & 10312                 & 23.23              & 22.34                      \\
Homology     & 12312                 & 24.14              & 26.0                    \\
 \bottomrule

\end{tabular}
\end{table}

\begin{table}[h!]
\centering
\caption{Restricted Data Setting: Fluorescence}
%\label{tab:my-table}
%\begin{tabular}{@{}lllll@{}}
\begin{tabular}{>{\centering\arraybackslash}m{5.0cm}|>{\centering\arraybackslash}m{2.5cm}|>{\centering\arraybackslash}m{2cm}|>{\centering\arraybackslash}m{2.5cm}>{\centering\arraybackslash}m{2cm}}

\toprule
Task    & Training Samples     & R2DL Test Accuracy & Bi-LSTM Test Accuracy \\ \midrule
Fluorescence     & 10769                 & 12.09              & 06.23                   \\
Fluorescence     & 25769                 & 34.26              & 37.93                     \\
Fluorescence     & 45769                 & 69.28              & 66.34                      \\
Fluorescence     & 53769                 & 66.34              & 68.0                    \\
 \bottomrule

\end{tabular}
\end{table}

\begin{table}[h!]
\centering
\caption{Restricted Data Setting: Solubility Prediction}
%\label{tab:my-table}
%\begin{tabular}{@{}lllll@{}}
\begin{tabular}{>{\centering\arraybackslash}m{5.0cm}|>{\centering\arraybackslash}m{2.5cm}|>{\centering\arraybackslash}m{2cm}|>{\centering\arraybackslash}m{2.5cm}>{\centering\arraybackslash}m{2cm}}

\toprule
Task    & Training Samples     & R2DL Test Accuracy & Bi-LSTM Test Accuracy \\ \midrule
Solubility     & 2500                 & 011.0              & 07.23                   \\
Solubility     & 4000                 & 47.26              & 39.93                     \\
Solubility     & 5200                 & 85.23              & 87.34                      \\
Solubility     & 6623                 & 94.0               & 93.1                    \\
 \bottomrule

\end{tabular}
\end{table}

% \pagebreak

% \subsection*{Algorithms}

% \begin{algorithm}[h!]
% \SetAlgoLined
% \KwInput{Pretrained sentiment classifier $X$, source model training data $\{V_S\}^{n}_{i=1}$, target model training data $\{V_T\}^{m}_{j=1}$, maximum number of iterations $T_1$ for updates to $\theta$, number of iterations $T_2$ for k-SVD, output-label mapping function $h$(·), step size $\{\alpha_i\}^{T_1}_{i=1}$}
% \KwOutput{Optimal adversarial program parameters $\theta$}
%  1. Random initialization of $\theta$\;
%  2. Define objective function for k-SVD from (1) \;
%  3. \While{$t_1 \leq T_1$}{
%       4. \While{$t_2 \leq T_2$}{ 
%         use approximate k-SVD to solve $V_T \approx  V_S \theta$ \; 
%         $t_2 \longleftarrow t_2 + 1$ \;
%     }
%   \textbf{\#Loss evaluation for $\theta$} \;  
% %  \If{Loss($\theta$) decreases}
%    $\theta \longleftarrow \theta - \alpha_{t_1} \cdot \nabla_\theta h(\mathbf{C}(f_\theta(X)) $\;
  
%   $t_1 \longleftarrow t_1 + 1$ \;
%  }
 
%  \caption{Representation Reprogramming via Dictionary Learning (R2DL)}
% \end{algorithm} 

\end{document}